\documentclass{article}
\usepackage[utf8]{inputenc}
\usepackage[a4paper, left=1.2in, right=1.2in, top=1.2in, bottom=1.2in]{geometry}
\usepackage{authblk}

\usepackage[numbers,sort]{natbib}
\usepackage{microtype}
\usepackage{graphicx}
\usepackage{subfigure}
\usepackage{booktabs}
\usepackage[hidelinks]{hyperref}
\usepackage{amsmath}
\usepackage{amssymb}
\usepackage{mathtools}
\usepackage{amsthm}
\usepackage{mathrsfs}
\usepackage{thmtools}
\usepackage{xspace}
\usepackage{xcolor}
\usepackage[parfill]{parskip}

\usepackage{multirow} 

\usepackage{listings}
\usepackage{xcolor} 
\definecolor{deepgreen}{rgb}{0.0, 0.5, 0.0}

\theoremstyle{plain}

\theoremstyle{definition}

\theoremstyle{remark}

\makeatletter
\DeclareRobustCommand\onedot{\futurelet\@let@token\@onedot}
\def\@onedot{\ifx\@let@token.\else.\null\fi\xspace}

\def\eg{\emph{e.g}\onedot}

\makeatother

\definecolor{cherry}{RGB}{128,0,32}
\newcommand{\cherry}[1]{{\textcolor{cherry}{#1}}}

\title{C-Procgen: Empowering Procgen with Controllable Contexts}

\author[1*]{Zhenxiong Tan}
\author[2*]{Kaixin Wang}
\author[1]{Xinchao Wang}

\affil[1]{National University of Sinagpore}
\affil[2]{Technion}

\date{\url{https://github.com/zxtan98/CProcgen}}

\begin{document}

\maketitle

\let\thefootnote\relax\footnotetext{* Equal contributions}

\begin{abstract}
We present C-Procgen, an enhanced suite of environments on top of the Procgen benchmark~\cite{cobbe2020leveraging}.
C-Procgen provides access to over 200 unique game contexts across 16 games.
It allows for detailed configuration of environments, ranging from game mechanics to agent attributes.
This makes the procedural generation process, previously a black-box in Procgen, more transparent and adaptable for various research needs.
The upgrade enhances dynamic context management and individualized assignments, while maintaining computational efficiency. 
C-Procgen's controllable contexts make it applicable in diverse reinforcement learning research areas, such as learning dynamics analysis, curriculum learning, and transfer learning. 
We believe that C-Procgen will fill a gap in the current literature and offer a valuable toolkit for future works.
\end{abstract}

\section{Introduction}
\label{intro}
To make reinforcement learning (RL) algorithms more robust, adaptable, and generalizable, the research community has built procedurally generated environments~\cite{cobbe2020leveraging, justesen2018illuminating, MinigridMiniworld23} to facilitate evaluating algorithms.
These environments, characterized by their ability to introduce variations within a single game using different contexts, require agents to avoid overfitting to any particular set of contexts~\cite{zhang2018dissection,zhang2018study,song2019observational,cobbe2019quantifying}. 
Among various environments available, Procgen~\cite{cobbe2020leveraging} is one of the most comprehensive large-scale benchmarks. 
With its 16 high-quality and diverse games, each complete with unique landscapes and challenges, Procgen has been widely used in the study of generalization and sample efficiency in reinforcement learning~\cite{kirk2023survey, wang2020improving, agarwal2021deep, laskin2020reinforcement, cobbe2020phasic}.

However, Procgen does not support precise control of the contexts.
The procedural generation of each level is a black-box process to users.
For example, in the game \cherry{\texttt{leaper}} (see Figure~\ref{fig:limitation} for a screenshot), users are not able to explicitly specify the number of lanes to create a curriculum.
This limits its full potential in domains like curriculum reinforcement learning~\cite{narvekar2020curriculum}, transfer reinforcement learning~\cite{zhu2023transfer}, and meta reinforcement learning~\cite{beck2023survey}, where precise manipulation of the environment’s contexts becomes necessary.
We will give a detailed discussion in Section~\ref{sec:procgen}.
While some simple environments with context control already exist, prior research has emphasized the demand for a large-scale, high-quality, game-like environment~\cite{romac2021teachmyagent,portelas2020automatic,kirk2023survey}.

To bridge this gap, we introduce C-Procgen to empower the original Procgen with explicit context control.
Specifically, C-Procgen encompasses all 16 games in Procgen, enhancing each game with a variety of adjustable context parameters.
These parameters span aspects such as game mechanics, agent attributes, map complexity, and game-specific features.
At the same time, C-Procgen retains the high simulation speed of Procen, incurring only a negligible overhead for switching context.
We believe that C-Procgen offers a more refined, flexible, and versatile environment for a wide range of research avenues in reinforcement learning.
To illustrate this, we highlight several intriguing future prospects that can harness the strengths of C-Procgen (Section~\ref{sec:usage_cprocgen}).



\section{What Hinders the Potential of Procgen?}
\label{sec:procgen}

\begin{figure}[t]
  \centering
  \includegraphics[width=0.95\textwidth]{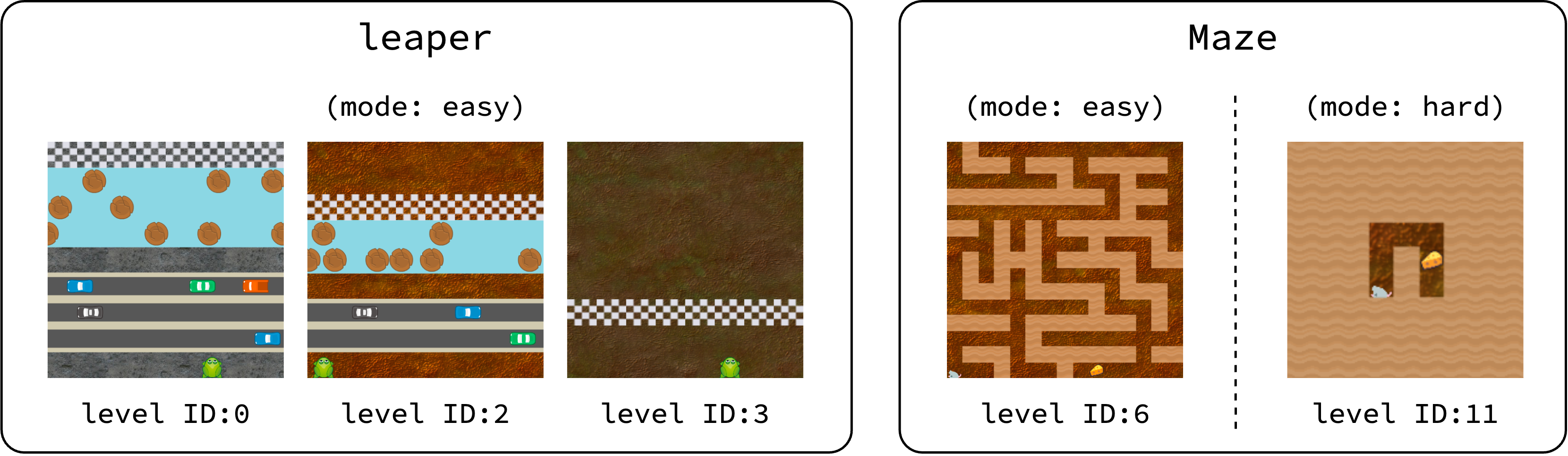}
  \caption{Procgen does not offer precise control over the game contexts, preventing us from generating the desired environments. \textbf{Left}: Different levels in the same \texttt{easy} mode can differ greatly in difficulty. \textbf{Right}: Sometimes, games in \texttt{easy} mode are even harder than those in \texttt{hard} mode.}
  \label{fig:limitation}
\end{figure}


In Procgen, the level generation process can only be controlled with the \texttt{difficulty mode} parameter in a somewhat coarse manner.
While the layout and landscape of a given level are uniquely determined by the \texttt{level ID}, this parameter is essentially a numerical identifier and lacks contextual information about the specifics of the level.
In the following, we will delve into a detailed discussion on the limitations of Procgen.

\textbf{Lack of Contextual Insight}

In Procgen, each episode starts in a new level procedurally generated with different contexts, unlike Atari games where the game's context remains constant.
Thus, the agent's behavior might vary considerably across different contexts.
However, Procgen does not offer a way to discern the contexts without hacking the low-level source code, rendering the environment as a black box.
If we had access to the detailed context information for each episode, we would be able to conduct a more comprehensive analysis of the agent's behavior.
For example, we could investigate how the agent adapts to various contextual factors, such as changes in the layout, obstacles, or objectives within the procedurally generated levels.
This level of insight could enable us to uncover strategies and decision-making patterns that the agent employs in response to different contexts. Additionally, having access to detailed context information could facilitate the development of more sophisticated and context-aware reinforcement learning algorithms, enhancing the agent's ability to generalize its skills across a wide range of scenarios.


\textbf{Coarse-Grained Control Over Game Context}

In addition to having no access to the underlying contexts, we are also constrained by very limited control over the contextual parameters in Procgen.
The only available options are to select between two difficulty modes: \texttt{easy} or \texttt{hard}.
The way these contexts are chosen to construct a new game level has been entirely hardcoded by the developers, which can significantly restrict Procgen's adaptability to various research requirements.
At times, this lack of fine-tuned control might lead to unexpected discrepancies.
For instance, levels may exhibit substantial variations in difficulty even when categorized under the same \texttt{easy} mode, as seen in Figure~\ref{fig:limitation} (left).
Furthermore, there may be instances where a level in the \texttt{hard} mode turns out to be much easier than one in the \texttt{easy} mode, as shown in Figure~\ref{fig:limitation} (right).
If we were granted more precise control over game contexts, we could define game distributions tailored to our specific research needs.
This would be particularly beneficial in settings like meta reinforcement learning, where we can systematically manipulate various context parameters and assess the agent's adaptability.


\textbf{Static Environment Parameters}

When using Procgen's parallel environments feature, each environment is constrained to the same \texttt{difficulty mode}.
Moreover, throughout the training process, the environment parameters remain static and unmodifiable.
Once the environments are initialized, we are unable to make any changes to the context configurations.
This limitation can be particularly problematic in the context of automatic curriculum learning~\cite{portelas2020automatic}, where the ability to dynamically adjust environment contexts based on agent performance is crucial for efficient training.
For example, researchers might want to gradually increase the difficulty of the environments by adjusting certain context parameters as agents demonstrate improved performance, thereby promoting more efficient learning.
However, due to the static nature of Procgen's environment parameters, such adaptive curriculum adjustments are not attainable.


\section{C-Procgen}
\label{cprocgen}

\begin{figure}[t]
  \centering
  \includegraphics[width=0.7\textwidth]{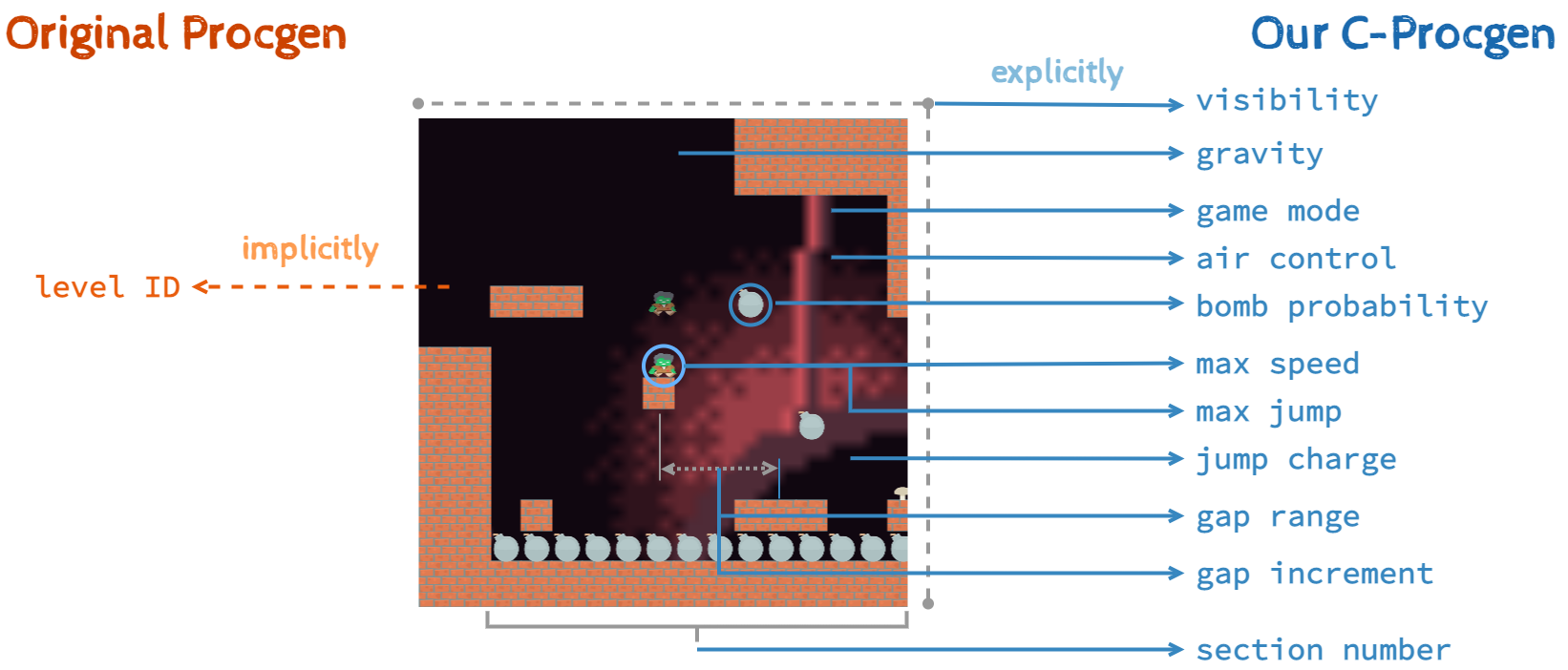}
  \caption{An illustrative example comparing the original Procgen and our C-Procgen.}
  \label{fig:ninjaIllu}
\end{figure}

We carefully refactor the source code of Procgen and expose many parameters that determine the context, essentially graying the black-box generation process.
The resulting benchmark is C-Procgen, which augments the original Procgen with \textbf{C}ontrollable \textbf{C}ontext parameters.

\textbf{Direct and Rich Contextual Parameter Configuration}

As illustrated in Fig \ref{fig:ninjaIllu}, the original Procgen confines users to a simple \texttt{level ID}, which offers only implicit control over the generation of game contexts.
In contrast, the enhanced version, C-Procgen, allows users to directly control the context via explicit configuration of over 200 unique contextual parameters across the suite of 16 Procgen games.
The diversity of these parameters encompasses:

\begin{enumerate}
    \item \textbf{Game Mechanics}: \eg, the maximum episode duration, the visible region size.
    \item \textbf{Reward Structures}: ones adjusting the intermediate reward values and the rewards upon game end.
    \item \textbf{Agent Attributes}: \eg, the agent's speed, health, and jumping capabilities.
    \item \textbf{Map Complexity}: ones allow users to set map dimensions, determine the number of sections, or configure maze complexities.
    \item \textbf{Game-Specific Features}: for instance, one can specify the initial size of the fish in \cherry{\texttt{bigfish}}, the variety of obstacles in \cherry{\texttt{cavefly}}, or the ball dimensions \cherry{\texttt{dodgeball}}.
\end{enumerate}

The flexibility and range of C-Procgen's contexts allow researchers to craft curricula within an expansive contextual space.
Furthermore, it is instrumental in some studies, such as edge-case testing and meta reinforcement learning, by facilitating the design of specific game settings to suit experimental needs.

\textbf{Individualized Context Assignments in Vectorized Setups}

In addition, C-Procgen also provides several engineering enhancements for improved usability. 
Unlike the original Procgen, where the same configuration is assigned to all environments during the initialization of vectorized environments, C-Procgen provides the method (Listing \ref{lst:python}) to assign distinct contexts to each environment. 
Such flexibility is useful in exposing the algorithm to richer game contexts during training.

\label{code:newCProcgen}
\begin{lstlisting}[caption={Defining and initializing game environments with specific contexts using C-Procgen.}, label=lst:python]
# Define two different contexts
context_1 = {
    "min_num_sections": 2,
    "max_num_sections": 6,
}
context_2 = {
    "air_control": 0.2,
    "visibility": 6,
}
# Create C-Procgen environments with specific contexts
env = CProcgenEnv(
    num_envs=2, env_name='ninja', 
    context_options=[context_1, context_2]
)
\end{lstlisting}

\textbf{Dynamic Context Management and Analysis}

Besides, one added convenience is the ability to modify the context of each environment between two episodes without instantiating a new environment, as shown in the `\texttt{set\_context\_to}' method (Listing \ref{lst:python2}). 
This feature facilitates the creation of a dynamic context distribution that could be of interest to the curriculum reinforcement learning community.

\begin{lstlisting}[caption={Assigning a new context to a running environment.}, label=lst:python2]
# Define a new context
new_context = {
    "min_num_sections": 1,
    "max_num_sections": 1,
}

# Assign the new context to the first environment
env.set_context_to(0, new_context)
\end{lstlisting}

Moreover, for games where the context varies within a specified range and is randomly generated for each episode, C-Procgen provides a method \texttt{env.get\_context()} to track the context for each episode.

\textbf{High Efficiency}

The enhancements brought by C-Procgen come with a minimal computation overhead, maintaining simulation efficiency similar to the original Procgen.
Table~\ref{tab:efficiency_comparison} provides a speed comparison.
For \texttt{env.step()}, the computation cost in C-Procgen is only 10\% higher than Procgen. 
The computational costs of the \texttt{env.get\_context()} and \texttt{env.set\_context\_to()} methods are approximately 1/5 and 2/3 of the original Procgen's \texttt{env.step()}, respectively. 
Considering that these two methods, in practical use, are generally executed only once per episode, the additional computational overhead they would introduce can be considered negligible.

\begin{table}
    \centering
    \begin{tabular}{lccc}
        \toprule
        \multirow{2}{*}{Method} & \multicolumn{3}{c}{Time Cost (ms)}  \\
        \cmidrule(l){2-4}
        & Procgen & C-Procgen & C/P \\
        \midrule
        \texttt{env.step()} & 109,934 & 121,575 & 110.6\% \\
        \texttt{env.get\_context()} & - & 23,992 & 21.8\% \\
        \texttt{env.set\_context\_to()} & - & 72,833 & 66.3\% \\
        \bottomrule
    \end{tabular}
    \caption{Computational cost comparison between C-Procgen and Procgen. Time costs for key methods are listed for both versions. The time costs presented in the table are measured under the condition of 64 environments running concurrently. Each function is executed 1,000,000 times in every environment. The ``C/P'' column shows the relative cost of C-Procgen methods as a percentage of the baseline `\texttt{env.step()}' from Procgen.}
    \label{tab:efficiency_comparison}
\end{table}

In summary, C-Procgen aims to provide more controllable context parameters and flexibility in game environment setups.
While offering explicit configurability and individualized assignments, it retains computational efficiency.
C-Procgen holds the promise of catalyzing progress in curriculum reinforcement learning research.

\section{How to Use C-Procgen}
\label{sec:usage_cprocgen}

The increased flexibility and fine-grained control offered by C-Procgen can serve as a very useful tool for reinforcement learning research.
Here are a few illustrative examples and potential avenues for harnessing the capabilities of C-Procgen:

\begin{enumerate}
    \item \textbf{Contextual Insights into Learning Dynamic}\\
    With the inclusion of context information, we can now approach the analysis of learning dynamics from the perspective of context. This provides a way to explore how various metrics, such as score, loss, and policy entropy, evolve as the agent learns across distinct contexts. Furthermore, analyzing episode lengths within different contexts during the learning process is significant, as it directly affects sample balance. Specifically, contexts with longer episode lengths contribute more samples, paving the way for a deeper understanding of learning dynamics in RL.
    
    \item \textbf{Curriculum Learning}\\
    The capability of C-Procgen to dynamically modify contexts proves useful in curriculum-based approaches~\cite{narvekar2020curriculum, portelas2020automatic}. It serves as a foundation for investigating the optimal selection and evolution of contexts that align with an agent's progressing abilities. Starting with basic configurations, the agent can understand the game's fundamental mechanics. As the agent advances, the complexity rises, ensuring that contexts remain suitably challenging and align with the agent's skill level.
    
    \item \textbf{Transfer Learning}\\
    C-Procgen allows for the training of agents in specific game contexts, followed by evaluating their adaptability when exposed to drastically different contexts. This method promotes the assessment of the adaptability of various learning algorithms when faced with changes in game mechanics or reward distribution.

    \item \textbf{Context-aware Reinforcement Learning}\\
    C-Procgen facilitates context-aware reinforcement learning~\cite{lee2020context, chen2021context}, where agents adjust their strategies based on the current context. This encourages the development of adaptive agents that can generalize across different scenarios, making their strategies more robust and versatile.
    
    \item \textbf{Enhanced Diversity for Stronger Agents}\\
    With C-Procgen, it's possible to introduce an array of contexts not present in the original Procgen. This facilitates the creation of an environment set more diverse than its precursor, promoting the training of more generalized and resilient agents.
    
    \item \textbf{Edge Case Analysis}\\
    The exhaustive parameter control provided by C-Procgen makes it viable to construct edge cases or rare scenarios. This capability is invaluable for thoroughly testing agents in non-standard conditions, evaluating their resilience and adaptability. Furthermore, it allows for the generation of rare yet crucial situations, potentially overlooked in the procedural generation of Procgen.
    
    \item \textbf{Environment Design Research}\\
    C-Procgen empowers researchers to accurately alter and examine specific game mechanics, agent attributes, and map intricacies. This facilitates focused studies on how different environmental elements influence agent learning, also providing a platform for designing and experimenting with novel game dynamics, ultimately enriching insights for refining RL environments.

\end{enumerate}

The aforementioned applications of C-Procgen represent only a few examples of its potential uses.
We believe future investigations will uncover its full potential.


\section{Limitations, Discussion and Future Prospects}

While C-Procgen brings improvements to procedural environment generation with its diverse functionalities, it's essential to recognize its challenges. In this section, we'll delve into its limitations, discuss inherent issues, and touch upon potential paths forward. 

\textbf{Parameter Overload}\\
While the introduction of over 200 unique contextual parameters allows for fine-grained control, it may inadvertently lead to a parameter overload. This increases the complexity and presents challenges in efficient environment configuration and management. Keeping a balance between  flexibility and usability is something we need to consider.

\textbf{Evaluation Challenges}\\
Given that C-Procgen offers much more flexibility in terms of environment configurations, comparing performance metrics across different studies could be challenging. Variations in parameter settings can lead to inconsistencies in benchmarks, making it difficult to compare algorithms in a standardized manner.

\textbf{Optimization Challenges}\\
With a vast parameter space, optimization algorithms may encounter extended computational times. Navigating this expansive parameter landscape could be a double-edged sword, making it challenging to pin down the optimal configurations promptly.

\textbf{Lack of Evaluation Protocols}\\
One of the clear limitations of our current work with C-Procgen is the absence of specific evaluation protocols. Although the platform offers extensive flexibility and configurability, it does not provide standardized benchmarks or evaluation criteria.

\textbf{Future Enhancements}\\
The trajectory of C-Procgen seems promising. Anticipated future enhancements might include mechanisms for automated optimal parameter detection or advanced context analytics and visualization tools, assisting researchers in understanding the effects of specific contexts on agent learning.

\bibliography{main}
\bibliographystyle{main}

\newpage
\appendix

\end{document}